# Subject Independent Emotion Recognition using EEG Signals Employing Attention Driven Neural Networks[Ψ]

Arjun[*a], Aniket Singh Rajpoot[*b] and Mahesh Raveendranatha Panicker[a,c]

[a]Electrical Engineering, Indian Institute of Technology (IIT) Palakkad, India.

[b]Computer Science and Engineering, Indian Institute of Technology (IIT) Palakkad, India

**Abstract**

Electroencephalogram (EEG) based emotional analysis has been employed in medical science, security and human-computer interaction with good success. In the recent past, deep learning-based approaches have significantly improved the classification accuracy when compared to classical signal processing and machine learning based frameworks. But most of them were subject-dependent studies which were not able to generalize on the subject-independent tasks due to the inter-subject variability present in EEG data. In this work, a novel deep learning framework capable of doing subject-independent emotion recognition is presented, consisting of two parts. First, an unsupervised Long Short-Term Memory (LSTM) with channel-attention autoencoder is proposed for getting a subject-invariant latent vector subspace i.e., intrinsic variables present in the EEG data of each individual. Secondly, a convolutional neural network (CNN) with attention framework is presented for performing the task of subject-independent emotion recognition on the encoded lower dimensional latent space representations obtained from the proposed LSTM with channel-attention autoencoder. With the attention mechanism, the proposed approach could highlight the significant time-segments of the EEG signal, which contributes to the emotion under consideration as validated by the results. The proposed approach has been validated using publicly available datasets for EEG signals such as DEAP dataset, SEED dataset and CHB-MIT dataset. With the proposed methodology, average subject independent accuracies of 65.9%, 69.5% for valence and arousal classification in the DEAP dataset and 76.7% for positive-negative classification in SEED dataset are obtained. Further for the CHB-MIT dataset, average subject independent accuracies of 69.1%, 67.6%, 72.3% for Pre-Ictal Vs Ictal, Inter-Ictal Vs Ictal, Pre-Ictal Vs Inter-Ictal classification are obtained, which are state-of-the-art to the best of our knowledge. The proposed end-to-end deep learning framework removes the requirement of different hand engineered features and provides a single comprehensive task agnostic EEG analysis tool capable of performing various kinds of EEG analysis on subject independent data.

## 1. Introduction

Emotions, which are at the core of who we are, are linked to our thoughts, decision-making abilities, and cognitive processes. Research on emotional states can improve existing brain computer interface (BCI) frameworks, which can then be used to implement treatments for disorders such as autism spectrum disorder (ASD), attention deficit hyperactivity disorder (ADHD), and anxiety disorder [1]-[3]. Recognition and interpretation of emotional states has always been a significant research subject in the fields of biological science, psychiatry, cognitive science, and brain-driven artificial intelligence because of these important applications [4]-[6].

Several methods have been developed for emotion recognition which includes the use of both physiological and non-physiological signals. Non-physiological signals include facial expressions, voice signals, body gestures while physiological signals include electroencephalogram (EEG), electrocardiogram (ECG) signals and many more. While using non-physiological signals is simple and does not require any special equipment, such signals may be forged and therefore are not considered to be a true reflection of one's emotional state. Physiological signals, on the other hand, are difficult to be manipulated and hence more suited to accurate emotion recognition tasks [7]. Among the many physiological signals used to recognize emotions, EEG has gotten a lot of attention as it has the capability of capturing the brain state in real time with a high time resolution [8], [9].

Various studies have been done in the past that have specifically handled emotion recognition through physiological signals and predominantly EEG as in [9]-[24]. Algorithms using power spectral density (PSD) [10]-[12], sample entropy (SE) [13], differential entropy (DE) [9], [56], wavelet transform (WT) [14] and several other methods have been used for extracting meaningful features from the EEG data. However, these model driven approaches try to approximate the EEG signals to a well-defined structured linear model and hence may not capture the non-linear statistics in the data. Majority of the state of the art models recently use machine learning based classifiers like [15]-[22], PSD features with Naive Bayes classifiers in [10], [11], PSD and statistical features with

[Ψ]This work was supported by Indian Institute of Technology (IIT) Palakkad, India.
[*]Equal Contribution
[c]Corresponding author
E-mail address: mahesh@iitpkd.ac.in (Mahesh Raveendranatha Panicker)

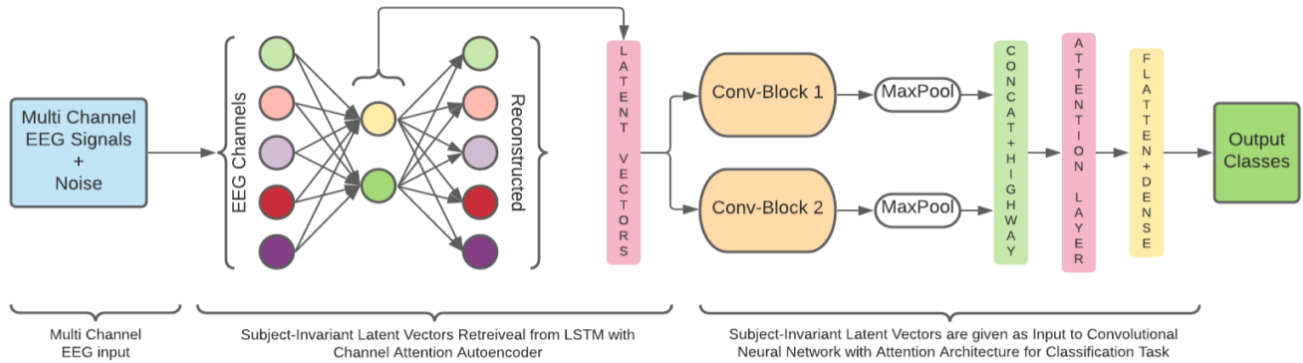

Fig. 1. The proposed framework for performing the subject-independent EEG emotion recognition. Where first the subject-invariant latent encodings are retrieved through the LSTM with channel-attention autoencoder separately and then are fed to the CNN with attn for classification task.

ontological models in [17], deep belief network (DBN) based features with support vector machine (SVM) in [18], power spectral and statistical features with neural networks in [19], features extracted using pyramidal convolutional neural network or autoencoders with fully connected networks as classifiers in [20]-[22]. Frequency domain-based feature extraction methods like short-time-Fourier transform (STFT) [23] and discrete wavelet transform (DWT) [14], [24] have also been used before.

The integration of signal processing and machine learning techniques has really helped in improving the results and the understanding of the emotion recognition task, but there still exists some major problems in most of the studies such as, using handcrafted features (leading to overfit) and poor generalization of the classifier model across subjects. Due to limitations of data in the medical field, it is difficult to get large, labeled dataset suitable for supervised learning techniques, thus the development of unsupervised learning techniques [25] is gaining significance. Thus, a new domain of subject-independent emotion recognition has been emerging recently where "leave one subject out" for validation is followed, a comprehensive review of the previous works in this domain can be found in Table 1. A more detailed review of the recent state-of-the-art works in both subject-dependent and independent emotion recognition domain can be found in [67]. In this research work we expanded on the results of previous studies in this area, based on the hypothesis that even among different individuals the basis of the neural networks formed in the brain while experiencing similar emotions remain the same [25]-[27].

One of the most significant components in any EEG analysis task is the accurate and compact low dimensional space (latent space) representation of the multichannel EEG signals. Autoencoders are found to be superior when compared with other nonlinear methods for finding the compact latent space representation of the given data. Generally other nonlinear dimensionality reduction methods rely on a single property like distance or topology which is not the case with autoencoders [28]. The overall proposed architecture is illustrated in Fig. 1. Attention mechanisms help in modeling dependencies without having the constraint of far distances in the sequence in contrast to CNN's which have localized networks and sequencing networks such as long short-term memory (LSTM) which do not have a good memory retaining capacity [29]-[32]. Therefore, our main contributions include the following aspects:

- A novel unsupervised LSTM with channel-attention autoencoder for getting the subject-invariant lower dimensional latent space representation of the EEG data where a similar representation space for different individuals is retrieved, thus alleviating the bottleneck created by the inter-subject variability of EEG data for subject-independent studies.
- A novel CNN with attention framework for performing the task of subject-independent EEG emotion recognition by combining the usage of autoencoder for getting the lower dimensional latent space representation of EEG data, the CNN with attention framework is also capable of identifying key time frames in the EEG signal in an unsupervised manner while the model learns to classify the signal under supervision.
- A thorough investigation of the temporal localization of emotional outbursts and of finding common latent vector subspaces irrespective of subjectivity of EEG data is performed.
- Investigation of the efficacy of the proposed model in a completely different classification setting (e.g., knowing the scope of the usage of the proposed architecture outside the emotion recognition domain). Towards this, the proposed model architecture is tested on the Children's Hospital Boston and the Massachusetts Institute of Technology Dataset (CHB-MIT) dataset [33] for epileptic seizure detection. As mentioned in some recent research papers [34], a seizure can be stimulated due to intense emotions and especially stress.

**TABLE 1**
**REVIEW OF PREVIOUS WORKS IN THE DOMAIN OF SUBJECT-INDEPENDENT EMOTION RECOGNITION WITH EEG**

| Research | Features | Classifier | Dataset Used |
|---|---|---|---|
| W.-C. L. Lew et al. [49] | Bi-GRU based features | Fully Connected Network (FCN) | DEAP [10] and SEED [35] dataset |
| Y. Luo et al. [53] | Differential Entropy features | Generative Adversarial Network | DEAP [10] and SEED [35] dataset |
| T. Song et al. [63] | Instance adaptive branch with Multi-level and Multi-graph convolutions | Long-Short-Term-Memory | SEED [35] and MPED [64] dataset |
| J. Chen et al. [54] | Statistical features | Ontological Model | DEAP [10] |
| V. Gupta et al. [58] | Information Potential(IP) features using FAWT decomposition | Random Forest | DEAP [10] and SEED [35] dataset |
| X. Li et al. [62] | Time-Frequency Domain Features and Non-linear dynamical systems features | Support Vector Machine | DEAP [10] and SEED [35] dataset |
| S. Liu et al. [65] | Dynamic Differential Entropy | Convolutional Neural Network | SEED [35] dataset |
| Z. Lan et al. [51] | Differential Entropy features | Logistic Regression Classifier | DEAP [10] and SEED [35] dataset |
| X. Li et al. [22] | Autoencoder based features | Long-Short-Term-Memory | DEAP [10] and SEED [35] dataset |
| J. Fernandez [66] | Power Spectral Density and Differential Entropy based features | Convolutional Neural Network with Stratified Normalization | SEED [35] dataset |

The rest of the paper is organized as follows: section two discusses the datasets employed and related methodology. The third section gives the comprehensive experimental results of the proposed algorithm and comparison algorithms followed by the fourth section on discussion and analysis of the proposed algorithm and results. The fifth section provides a conclusion to the paper.

## 2. Materials and Methods

### 2.1 Dataset

The proposed method was validated with the publicly available and widely used Database for Emotion Analysis using Physiological Signals (DEAP) [10], SJTU Emotion EEG Dataset (SEED) [35], and the CHB-MIT dataset [33]. The details regarding the number of subjects, nature of signals, duration and sampling rate of the EEG signal are presented in Table 2.

### 2.1.1 Data set 1: DEAP Dataset

The DEAP dataset [10], acquired from 32 subjects, was recorded simultaneously while each participant watched 40 handpicked one-minute music videos. The EEG recordings were taken originally via 32 channels at a sampling rate of 512 Hz which was later down sampled to 128 Hz and band pass filtered to 4-45 Hz. The recorded data was finally segmented into 60 seconds with the initial 3 seconds baseline data being removed. Each video was rated by the participants based on Arousal, Valence, Dominance, Liking and Familiarity. For Arousal, Valence, Dominance and Liking, rating was done on a continuous scale of 1-9 and for Familiarity it was done on a discrete scale of 1-5. With the DEAP dataset, many classes can be extracted from labels by dividing them equally. In the proposed work, the Valence and Arousal labels were used with each of them being equally divided into 2 classes i.e., <5 and >=5.

### 2.1.2 Data set 2: SEED Dataset

The SEED dataset [35] consists of EEG signals recorded from 15 participants (7 males and 8 females) who volunteered for the experiment thrice at an approximate interval of a week. The EEG data was recorded simultaneously while each participant watched 15 handpicked approximately four-minute Chinese movie clips. The EEG recordings were taken originally via 62 channels, placed according to the 10-20 system, at a sampling rate of 1000 Hz which was later down sampled to 200 Hz and band pass filtered to 0-75 Hz. In this dataset the labels are predefined, and, in this work, we will be using positive and negative labels.

### 2.1.3 Data set 3: CHB-MIT Dataset

**TABLE 2**
**SPECIFICATIONS OF THE DATASET**

|  | DEAP [10] | SEED [35] | CHB-MIT [33] |
|---|---|---|---|
| **No. of subjects** | 32 | 15 | 23 |
| **Nature of signals** | Various Physiological Signals | EEG | EEG |
| **No. of channels** | 32 | 62 | 22 |
| **Stimuli** | Music Videos | Movie Clips | No external stimuli |
| **Trials Per Session** | 40 | 15 | Several |
| **Trial Length** | 63 seconds | Approx. 200-240 seconds | 1 hour |
| **Sampling rate of EEG** | 128 Hz | 200 Hz | 256 Hz |

The CHB-MIT Dataset [33] consists of EEG signals recorded from 23 children (17 females and 5 males) with epilepsy. The EEG data was recorded for the children for up to several days following their withdrawal of anti-seizure treatment. The EEG recordings were taken originally via majorly 22 channels at a sampling rate of 256 Hz. Seizure's start and end time was provided in the dataset, which was determined by the experts. In the proposed work we will be classifying the Pre-Ictal Vs Ictal, Inter-Ictal Vs Ictal, Pre-Ictal Vs Inter-Ictal signals. Ictal is the period when the patient is going through a seizure, Interictal is the normal state between two seizures and preictal is the period when the transition occurs between interictal and ictal.

*2.2 Methods*

The proposed methodology as shown in Fig. 1 has three components: encoder, attention mechanism and a classifier. The details of the individual components are described below.

*2.2.1 Attention Mechanism*

Attention mechanisms have gained popularity in various deep learning tasks recently, ranging from sequence-sequence modeling to image-based tasks [31], [32]. Attention mechanism enables neural networks to perform well on multiple tasks irrespective of the input's length, size, or structure. In this work, attention mechanism is integrated into the auto-encoder architecture, where the work of attention was to determine where to apply focus on the vectors, encoded by the encoder such that the desirable output can be generated. The concept of attention mechanism is applied to the classifier head also. A typical attention layer is characterized by (1-3). The scalar weights $\beta_t$ as given in (1), is obtained as a dot product of the previous hidden state $h_t$ and the attention weight vector $W$. A SoftMax operation is applied on the scalar weights as shown in (2), where $T$ represents the total length of the input vector. Finally, the context vector $C$ is obtained by scaling all the hidden states over the output of the SoftMax operation, $\alpha_t$ and summing them up as shown in (3).

$$\boldsymbol{\beta_t} = \boldsymbol{h_t} \odot \boldsymbol{W} \tag{1}$$

$$\boldsymbol{\alpha_t} = \frac{e^{\beta_t}}{\sum_{j=1}^{T} e^{\beta_j}} \tag{2}$$

$$\boldsymbol{C} = \sum_{t=1}^{T} \alpha_t * h_t \tag{3}$$

*2.2.2 Latent Representation of EEG Channels*

The first step towards any classifier employing multi-channel EEG signals is a compact low dimensional representation of the high dimensional temporal signal. In the recent past, neural network-based approaches, particularly auto encoders due to its unsupervised learning nature, have been extremely popular for the latent space encoding of EEG signals. In this work, various methods for encoding the channels of EEG in a subject-invariant lower dimensional latent space using variations of autoencoders are experimented with a goal of achieving a similar representation space for even different subjects.

*(1) Autoencoder*

Autoencoder is an unsupervised learning technique which uses neural networks to learn the underlying feature space of the given dataset [36] and illustrated using Fig. 2 (a). Autoencoders consists of encoder and decoder blocks, in which the encoder learns to map the correlated input $(x_i)$ $x_i \in R^{d_x}$, to a latent space representation $h_j$ (4), and the decoder learns to map this latent space representation back to the input, $\overline{x^{(t)}}$, (5). This network is trained by minimizing the reconstruction loss $L(x, \overline{x})$, which is generally,

$$\boldsymbol{h_j}^{(t)} = f(\boldsymbol{W_j^1} \otimes \boldsymbol{x^{(t)}} + \boldsymbol{b_1}) \tag{4}$$

$$\overline{\boldsymbol{x^{(t)}}} = f\left(\boldsymbol{W_j^2} \otimes \boldsymbol{h_j^{(t)}} + \boldsymbol{b_2}\right) \tag{5}$$

$$\boldsymbol{L(W^1, W^2, b^1, b^2 : \chi)} = \sum_{x^{(t)} \in \chi} ||x^{(t)} - \overline{x^{(t)}}|| \tag{6}$$

the difference between the input $(x)$ and the reconstructed input $(\overline{x})$ by the autoencoder (6).

*(2) LSTM with Channel Attention Autoencoder*

The idea for LSTM autoencoders comes from the sequence-to-sequence models, which are typically used in text and time series-based tasks [37]-[40]. The working of LSTM revolves around the various gates that are inside it i.e., forget gate $(f_t)$, input gate $(i_t)$, cell gate $(c_t)$ and output gate $(o_t)$. In this work, we have used an LSTM with attention autoencoder architecture. LSTM's ability of taking the sequence and then encode the features of the data ensuring the temporal dependency, helps in better encoding of the data. This is because, in LSTM, the hidden states and cell states from the previous timesteps are shared with the current timestep as shown in (7).

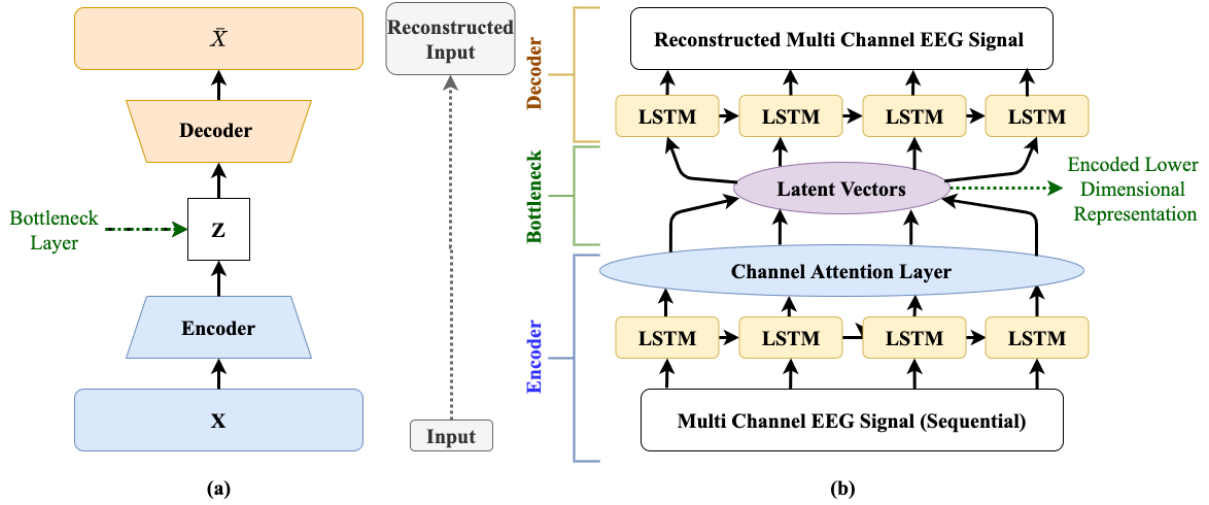

Fig. 2. (a) Typical Autoencoder, (b) Proposed LSTM with Channel-Attention Autoencoder.

$$h_t = \text{LSTM}(x_t, h_{t-1}, c_{t-1}) \tag{7}$$

In this work, to make the encodings by the LSTM autoencoder more robust, an attention layer [31], [32] is proposed in the encoder block. The attention is applied on the channel dimension of the EEG data, which helps in placing similar channel encodings closer in the latent space. The added attention layer also facilitates the decoding process and helps in giving better reconstruction of the signal input ($x_t$). For maintaining the timesteps dimension, channel-attention was applied such that the output of the SoftMax was just scaled over all the hidden states and the summation step was omitted in contrast to (3). The architecture of the proposed approach is as shown in Fig. 2 (b).

For increasing the feature extraction capability of the autoencoder, additive white Gaussian noise (AWGN) was added to the input, such that signal-to-noise ratio becomes unity. The autoencoder is trained such that it takes noisy data as the input and generates (decodes) the clean data as the output [41]. In this way, the encoder-decoder architecture can be made more robust and generalizable.

### 2.2.3 Classifier

Once the compact subject-invariant low dimensional channel agnostic latent space representation of the EEG signal is available, the proposed classifier for doing the classification task is applied on the same. The proposed architecture is the combination of embeddings from language models (ELMo) [47] inspired character CNN with a highway network [48] and an additional attention mechanism at the end which will summarize the temporal direction (time series axis), as illustrated in Fig. 3. A fully connected layer is added at the end for getting the number of classification classes. The CNN is formed by concatenating two convolutional blocks each capturing different aspects of the data i.e., the first convolutional block is designed such that it captures the nearby relationships, and the second convolutional block is designed such that it captures the long-range relationships. The first convolutional block contains a single convolutional layer, and the second convolutional block contains two convolutional layers, each convolutional layer (8) is followed by an instance normalization layer [43] (Eq. 9, where $\mu_{ic}$ and $\sigma_{ic}$ are the mean and standard deviation of the data along batch $i$ and channel c), a parametric rectified linear unit (PReLU) activation [44] (10), and a dropout layer. Before concatenating the two convolutional blocks they are first passed through a max pooling layer. To ease the gradient-based training of deep neural networks the convolutional features are passed through a highway network architecture [48] which learn to regulate the flow of information through a network by using gating units, as shown in Fig. 4. Finally, for getting a summarization of the time-axis, half of the features ($\widehat{h_{t_1}}$) from the highway network output is passed through a SoftMax activation layer (11) which acts as attention mechanism for the rest of the features [31], [32], [42] (12) as detailed in Fig. 3.

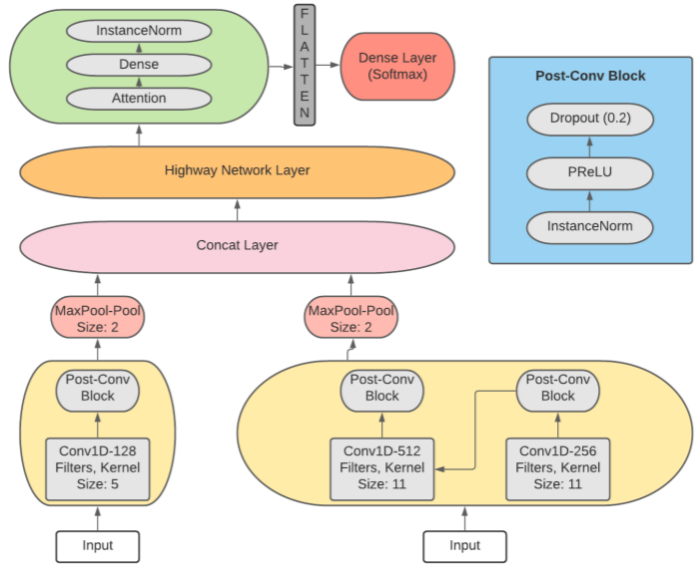

Fig. 3. The proposed CNN with Attention Architecture for the classification task.

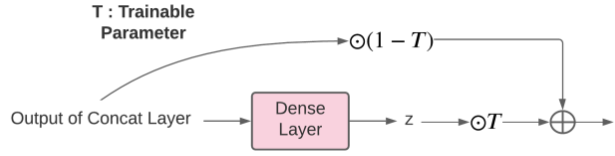

Fig. 4. The highway network used in the proposed CNN with Attention architecture.

$$y = W \otimes x + b \quad (8)$$

$$s_{icm} = \frac{y_{icm} - \mu_{ic}}{\sqrt{\sigma_{ic}^2 + \epsilon}} \quad (9)$$

$$h_t = s \ if \ s \geq 0, else$$
$$h_t = \eta s \ if \ s < 0, \eta = learnable\ parameter \quad (10)$$

$$a = \frac{e^{\widehat{h_{t_1}}}}{\sum_{j=1}^{T} e^{\widehat{h_{j_1}}}} \quad (11)$$

$$h = \widehat{h_{t_2}} \cdot a \quad (12)$$

Where, $\widehat{h_{t_2}}$ is the second half of the features from the highway network output and $a$ is the SoftMax of the first half of the features ($\widehat{h_{t_1}}$). The result of this attention mechanism is then passed through a dense layer followed by an instance normalization layer. Even though ablation studies have been conducted with various types of normalization and activation for this architecture, the instance normalization and PReLU activation was found to facilitate the training in the best manner.

## 3. Experiment and Results

In this section, an exhaustive validation of the proposed architecture is presented. A thorough comparison is done for both the components of the proposed architecture i.e.,
- The autoencoder approach is compared against other data driven encoding methods like principal component analysis (PCA), independent component analysis (ICA), and
- The proposed CNN-attention based classifier is compared against time series classifiers based on ResNet [45], Inception [46], LSTM [39] and Transformers [29], [30].

The proposed architecture is also compared with the continuous wavelet and Mel-spectrogram based image classification methods using a pre-trained VGG-16 [50]. The main reason for using images obtained from Mel-spectrogram in this work is due to the success of Mel-spectrogram in the audio-based tasks and since the dataset used here were taken through audio-video stimuli, Mel-spectrogram based images might be useful. In the end, the proposed approach is also compared with the state-of-the-art approaches.

*3.1 Experimental Setup*

To ensure the subject independent accuracy on the specified classes, the training is done in "leave-one subject out" manner, such that one subject is reserved for evaluation while the rest are used for training. The process is repeated for each subject in the dataset and the average was calculated to determine the subject independent accuracy. The implementation was done using Python 3.7.10 and TensorFlow 2.5.0, and both the proposed models i.e., LSTM with channel-attention autoencoder and CNN with attention, are trained using ADAM optimizer with a learning rate of 0.001. The complete architectural details and parameter choices of the LSTM with channel-attention autoencoder and CNN with attention classifier for different datasets are presented in Table S5-S8 as part of the supplementary material.

TABLE 3
VALIDATION RESULTS FOR DEAP DATASET WITH DIFFERENT ENCODR AND CLASSIFER ARCHITECTURES

| CLASSIFER | ENCODER | Autoencoder (AE) | LSTM AE with Attn | Variational-AE | PCA | ICA | CWT (coif5) | Mel-spectrogram |
|---|---|---|---|---|---|---|---|---|
| CNN with Attn | Valence | 62.6 | **65.9** | 63.7 | 61.3 | 59.7 | - | - |
|  | Arousal | 65.6 | **69.5** | 66.4 | 62.5 | 62.1 | - | - |
| LSTM | Valence | 60.3 | 61.7 | 59.3 | 57.7 | 56.2 | - | - |
|  | Arousal | 61.1 | 62.2 | 61.5 | 57.1 | 56.4 | - | - |
| ResNet 1D | Valence | 55 | 56.2 | 53.7 | 54.5 | 52.6 | - | - |
|  | Arousal | 55.7 | 56.5 | 56.2 | 55.6 | 54.5 | - | - |
| Inception 1D | Valence | 54.7 | 56.7 | 53.2 | 54 | 55.7 | - | - |
|  | Arousal | 55.5 | 55.9 | 55.4 | 56.1 | 53.2 | - | - |
| Transformer | Valence | 56.4 | 52.1 | 53.2 | 52.9 | 53.6 | - | - |
|  | Arousal | 55.2 | 55 | 54.7 | 54.3 | 55.3 | - | - |
| VGG-16 | Valence | - | - | - | - | - | 52.5 | 55.7 |
|  | Arousal | - | - | - | - | - | 56.3 | 55.3 |

As discussed, the first step is encoding of the data to obtain the subject-invariant lower dimensionality representation. Based on ablation studies, 8 is set as the number of latent dimensions to be obtained from the encoding methods for DEAP dataset (32 input EEG channels) and CHB-MIT (Refer to Table 5 for details on input channels) dataset and for the SEED dataset(62 input EEG channels), the latent dimensions to be obtained from the encoding methods is set as 16. Tables S3 and S4 in the supplementary material present the results of the proposed framework on different choices of latent dimensions. After getting the subject-invariant lower dimensional representation of the data, it is provided as the input to the proposed CNN with attention architecture for the classification task according to the dataset provided.

*3.2 Results*

*3.2.1 Data set 1: DEAP Dataset*

In most of the approaches using DEAP dataset, emotional states, which are preferably discretized into various states such as excitement, anxiety, rage, pleasure, surprise, and so on, are broadly divided into two approximate dimensions: valence and arousal. The valence variable determines the emotion's positive or negative effects, while the arousal dimension determines its intensity. The results obtained on the DEAP dataset by experimenting with various combinations of encoder and classifier architectures is shown in Table 3. The combination of the proposed LSTM with channel-attention autoencoder and CNN with attention architecture outperformed all other encoder + classifier frameworks. The results on combinations of encoder + classifier frameworks are presented in supplementary material Table S1. Furthermore, the idea of using CWT or Mel-Spectrogram based images along with a pretrained VGG-16 classifier did not yield good results.

*3.2.2 Data set 2: SEED Dataset*

The data for the SEED dataset has irregular recordings with recording duration ranging from 200-240 seconds approximately. Hence to ensure the same amount of data from each subject, it was decided to use the central 150 seconds of data from each individual's recording. Lastly, as in the SEED dataset there are 3 recordings for each individual recorded during different sessions. So, for covering all the sessions, the model was trained for each session separately and then the results were averaged for the 3 sessions. The results obtained on the SEED dataset by experimenting with various combinations of encoder and classifier architectures is shown in Table 4 and the superior nature of the proposed LSTM with channel-attention autoencoder and CNN with attention framework can be observe. Table S2 in the supplementary material has more results on combination of encoder + classifier frameworks.

TABLE 4
VALIDATION RESULTS FOR SEED DATASET WITH DIFFERENT ENCODR AND CLASSIFER ARCHITECTURES

| CLASSIFER | ENCODER | Autoencoder (AE) | LSTM AE with Attn | Variational-AE | PCA | ICA | CWT (coif5) | Mel-spectrogram |
|---|---|---|---|---|---|---|---|---|
| CNN With Attn | | 72.1 | **76.7** | 72.7 | 65.1 | 62.3 | - | - |
| LSTM | | 61.8 | 62.2 | 60.5 | 59.1 | 59.7 | - | - |
| ResNet 1D | | 58.9 | 59.5 | 59.1 | 57.1 | 56.8 | - | - |
| Inception 1D | | 57.5 | 59.9 | 59.4 | 56.1 | 56.3 | - | - |
| Transformer | | 58.1 | 60.1 | 59.2 | 58.7 | 59 | - | - |
| VGG-16 | | - | - | - | - | - | 59.3 | 61.1 |

TABLE 5
SELECTION CRITERIA FOR CHB-MIT DATASET

| Classification Task | No. of patients selected | Channels selected |
|---|---|---|
| Ictal vs Pre-Ictal | 11 | 22 |
| Ictal vs Inter-Ictal | 9 | 18 |
| Pre vs Inter-Ictal | 9 | 18 |

TABLE 6
VALIDATION RESULTS FOR CHB-MIT DATASET

| Classification Task | Proposed Framework |
|---|---|
| Ictal vs Pre-Ictal | 69.1 |
| Ictal vs Inter-Ictal | 67.6 |
| Pre vs Inter-Ictal | 72.3 |

*3.2.3 Data set 3: CHB-MIT Dataset*

The CHB-MIT dataset has recordings of the epileptic patients recorded over a long period of time. Hence, for doing the analysis on the proposed model, several 20 second segments of the recordings of the patient were chosen while being in the pre-ictal, interictal and ictal states. To ensure that enough continuous data is there from each epileptic state of each individual, a few patients were only selected from the 23 patients recording for each classification task. As the number of channels is also irregular in this dataset, the intersection of all the selected patient's channels was considered. The details for this can be seen in Table 5. The results of the CHB-MIT dataset using the proposed framework (LSTM with channel-attention autoencoder and CNN with attention) can be seen in Table 6.

*3.3 Comparison with Approaches in Literature*

The proposed approach is also compared in a comprehensive fashion with well-established approaches in literature and the results are reported in Table 7. From Table 7, it can be seen that, the proposed LSTM with channel-attention autoencoder and CNN with attention architecture performs at par with the recent-related state-of-the-art studies documented in literature and in fact achieves state of the art performance on the CHB-MIT Dataset for epileptic seizure state detection to the best of our knowledge. The main reason for getting good results through the proposed architecture could be attributed to the attention-based mechanism. First, we used the LSTM with channel-attention autoencoder for getting the subject-invariant lower dimensional latent space representation of the EEG data, which helps in getting a closer latent vector subspaces for different individuals. Then we used the CNN with attention classifier for classifying the time-series signals, in which the attention mechanism plays an important role for identifying the important time segments with respect to the classification task. More detailed analysis of these aspects is presented in the following section and in the Discussion section.

3.4 *Latent Time Analysis*

To determine the exact instance of the onset of emotion after applying a stimulus, an analysis of the trained attention weights from the CNN with attention framework used for classification has been done. On analyzing the activation values for the DEAP dataset, we found a huge difference in the activation values of the neural network before and after the attention layer. It was interesting to note that, after the attention layer, the regions of high activation for each individual occurred over short time bursts. This observation aligns with the findings in emotion research suggesting that emotions are not continually represented in the brain but , ut outbursts of activations take place over a few short time segments [57]. The visualization of these findings for the DEAP dataset can be seen in Fig. 5. To the best of our knowledge, this is the first attempt to localize the emotion to exact time duration in the EEG signal computationally.

Furthermore, we also analyzed the trained attention weights from the CNN classifier trained for recognition of an ictal epileptic state in the EEG data provided by the CHB-MIT dataset. The EEG input signal consisting of the three epileptic stages and the corresponding activation values is shown in Fig. 6. As can be seen in the figure, on the onset of the ictal state the activation values of the neurons start firing up. So, from this observation we can say that the proposed CNN with attention classifier learns to correctly identify the important time segments in the EEG signal in an unsupervised manner while the model learns to classify the signal under supervision. Thus, this makes the proposed CNN with attention classifier a very potent tool for classifying EEG signals along with also identifying the key time frames responsible for enabling the particular classification.

## TABLE 7
## COMPARISON WITH STATE-OF-THE-ART APPROACHES
### DEAP DATASET

| Research | Features | Classifier | Valence | Arousal |
|---|---|---|---|---|
| W.-C. L. Lew et al. [49] | Bi-GRU based features | Fully Connected Network (FCN) | 56.8 ± 3.3 | 56.6 ± 3.5 |
| Y. Luo et al. [53] | Differential Entropy features | Generative Adversarial Network | 68 ± 6.6 | 66.8 ± 5.5 |
| J. Chen et al. [54] | Statistical features | Ontological Model | 67.8 | 68.96 |
| V. Gupta et al. [58] | IP features using FAWT decomposition | Random Forest | 71.43 | 79.9 |
| **Ours** | **LSTM with Channel Attention Autoencoder** | **CNN with Attention** | **65.9 ± 9.5** | **69.5 ± 9.7** |

### SEED DATASET

| Research | Features | Classifier | Accuracy (Pos-Neg) |
|---|---|---|---|
| Z. Lan et al. [51] | Differential Entropy features | Logistic Regression Classifier | 72.47 ± 12.47 |
| X. Li et al. [22] | Autoencoder based features | LSTM | 84.2 |
| V. Gupta et al. [58] | IP features using FAWT decomposition | Random Forest | 90.48 |
| **Ours** | **LSTM with Channel Attention Autoencoder** | **CNN with Attention** | **76.7 ± 8.5** |

### CHB-MIT DATASET

| Research | Features | Classifier | Ictal Vs Pre-Ictal | Ictal Vs Inter-Ictal | Pre vs Inter-Ictal |
|---|---|---|---|---|---|
| T. Dissanayake et al. [52] | CNN based features | Transfer Learning with CNN | - | - | 55.84 |
| **Ours** | **LSTM with Channel Attention Autoencoder** | **CNN with Attention** | **69.1 ± 11.9** | **67.6 ± 10.6** | **72.3 ± 11.2** |

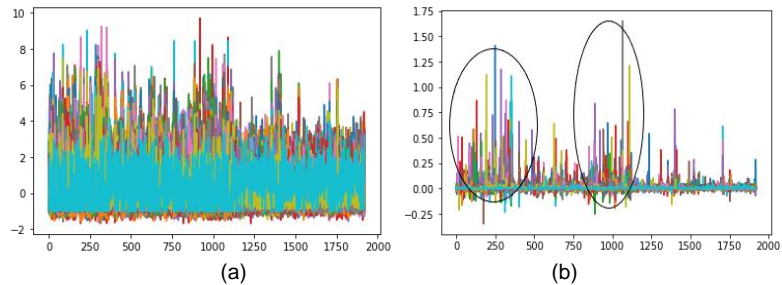

Fig. 5. On the left we can see the activation values of all the filters before the attention layer for an individual, where each color represents the activation value of a separate filter, (a) and on the right we can see the activation values of all the filters after the attention layer for an individual, (b). x-axis: time axis compressed by CNN to 1920 data points representing 60 seconds, y-axis: activation values of CNN nodes.

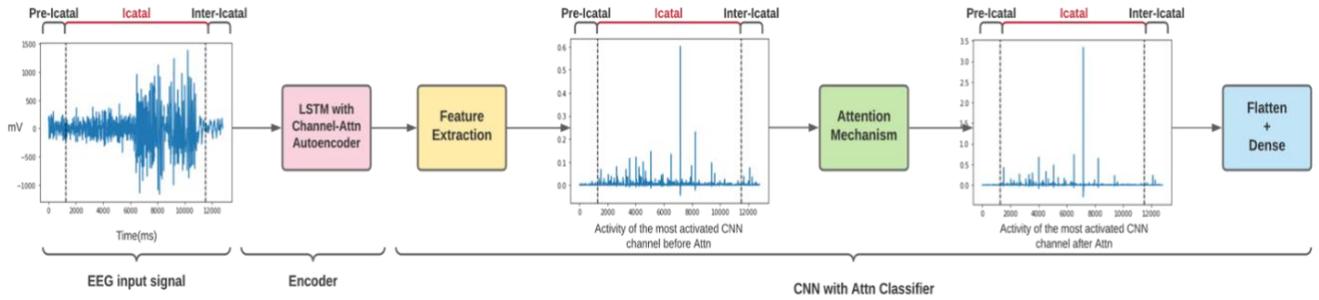

Fig. 6. An illustration of how the proposed framework can be successfully employed for identifying the onset of an epileptic seizure. As can be clearly seen in the plot of activity values of the most activated CNN channel after attention, the activity of neurons starts firing on the onset of the Ictal state. The ground truth onset of different epileptic states is marked with the dotted vertical lines based on the annotations provided in the CHB-MIT dataset

*3.5 Latent Dimension Analysis*

To identify the reason for enhanced performance of classifier models using lower dimensional latent space representation of the EEG data obtained from LSTM with channel-attention autoencoder, the t-Distributed Stochastic Neighbor Embedding (t-SNE) [55] was employed for the latent space analysis. These techniques in a way help in dimensionality reduction for convenient and better visualization. It is well known that there is a significant variation between the patterns of EEG data obtained from different individuals while experiencing the same emotion. This subjectivity and variability in EEG data often lead to failure of generalization on data even by highly capable architectures. To understand this observation better, dimensionality reduction to 8 dimensions is performed on the 32-channel raw-EEG data provided in the DEAP dataset by using PCA and our proposed LSTM with channel-attention autoencoder for a few individuals. The results are illustrated in Fig. 7, where each color represents the lower dimensional representation of an individual's data. The separation in the distribution of the lower dimensional representations for each individual is clearly evident from the PCA obtained plot, Fig. 7 (a). This also conveys that each individual exhibits or manifests emotion in a vividly different manner and thus the intrinsic variables in it cannot be captured using dimensionality reduction methods like PCA. In contrast to PCA, the lower dimensional representations obtained from the proposed LSTM with channel-attention autoencoder Fig. 7 (b). consists of much closer latent vector subspaces for different individuals, which symbolizes that the autoencoder was able to capture the intrinsic variables present in the EEG data. This observation shows that the proposed LSTM with channel-attention autoencoder is capable of learning subject-invariant latent vector subspaces i.e., intrinsic variables, for EEG data belonging to different individuals and break the barriers created by the subjectivity of EEG data. Thus, this provides us with a useful tool for obtaining lower dimensional representation of EEG data on which any downstream task can be performed for analyzing the EEG data in a subject-independent manner.

3.6 *Localization in Space (EEG Channels) Analysis*

It can be said that each human individual perceives emotions uniquely and the pattern of activity in different brain regions can be different too, as evident from Fig. 8, where each individual is experiencing a happy emotion. So, we tried to investigate whether this different localization of EEG channels is preserved by the proposed LSTM with channel-attention autoencoder for different individuals. To do so, the most similar raw EEG channels are identified for each latent dimension obtained from the LSTM with channel-attention autoencoder using cosine similarity. It is observed that the highly activated latent dimensions were generally close to the highly activated raw EEG channels and the same was observed for the low activated regions. Thus, the conservation of localization of EEG channels for each individual in the latent dimensions obtained from the LSTM with channel-attention autoencoder is observed, but these observations need to be thoroughly validated through a larger dataset and various other

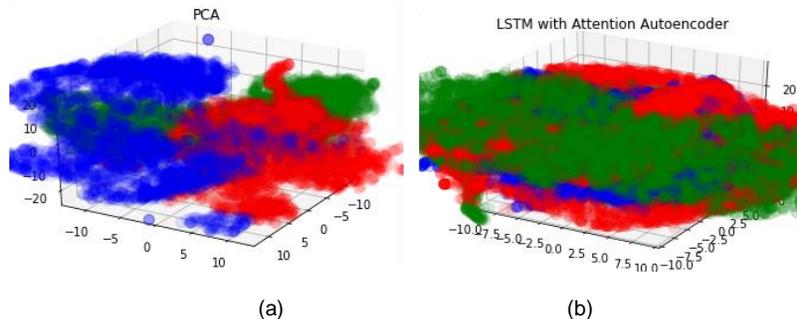

(a)      (b)

Fig. 7. Red, Blue, and Green data points correspond to the lower dimensional representation of 3 different individual's data. (a) The t-SNE plot of 3 different subjects using the 8-dimensional latent representation obtained from PCA. (b) The t-SNE plot of 3 different subjects using the 8-dimensional latent representation obtained from LSTM with channel attention autoencoder.

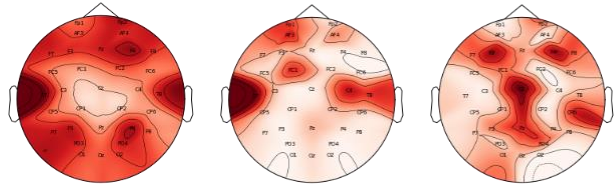

Fig. 8. The EEG electrode placement map plot is shown for 3 different individuals from DEAP dataset. As can be seen in different individual's different parts of the brain is activated while "*happy*" emotion formation.

sophisticated methods, which will be conducted in a future study.

## 4  Discussion

An end-to-end deep learning framework was proposed which first utilizes an LSTM with channel-attention autoencoder for encoding the input EEG signal such that a similar latent vector subspace is obtained for even different individuals as was seen in section 3.5. Then a CNN with attention classifier was used to classify the subject-invariant encodings obtained from the LSTM with channel-attention autoencoder into different labels for tasks like emotion recognition and epileptic seizure detection. Also, an approach to identify the important time segments in the EEG signal in an unsupervised manner was attempted as seen in section 3.6. Moreover, to show the characteristic of the proposed end-to-end deep learning framework to be application agnostic, it was trained to solve the task of epileptic seizure detection. To the best of our knowledge this was the first attempt towards showing application agnostic deep neural networks for EEG signals and the state-of-the-art results have been achieved on the CHB-MIT dataset with respect to subject-independent accuracy.

As can be seen from Table 7, most of the approaches in previous works [51-54], [58] heavily rely on hand-engineered features like differential entropy, statistical features, information potential and FAWT decompositions. These hand-engineered features can be very sensitive to the dynamics of the EEG signal such that the hand-engineered features which worked out for one task won't guarantee success with other tasks as the dynamics of EEG signals vary widely between different tasks. And, with the growing usage of automated systems for analysis of EEG signals it can become inefficient to find the appropriate signal-processing/statistical analysis technique for feature extraction for every signal task. Thus, by the usage of deep learning-based architectures we can alleviate this problem as task dependent features can be learnt while training. But, in the case of "leave-one subject out", subject-independent EEG analysis, even the deep learning-based architectures can fail as the distribution of EEG data for each individual can form a separate cluster in the space as seen in Fig. 7 (a). To solve this problem, caused due to inter-subject variance of EEG data, we proposed a LSTM with channel-attention autoencoder capable of obtaining a subject-invariant latent vector subspace encoding for different individuals. The application of a LSTM with channel-attention type autoencoder is not only limited to retrieving subject-independent encodings but also has potential use in many neuroscience applications like analyzing EEG recordings from individuals suffering from diseases like Alzheimer's or Schizophrenia [59-61], where basic functional brain networks may have been disrupted. Furthermore, by the usage of emerging techniques for the analysis of the weights and activation values of the deep learning models we can attain more insights to the behavior of the input EEG signal with respect to time i.e., neuronal dynamics and investigate applications like identifying key timeframes with respect to the task being solved as showcased in section 3.6. Such applications can tremendously help clinicians by directly guiding them towards the important timeframes of lengthy EEG recording for further investigation.

The proposed deep learning-based frameworks are relatively compact with respect to other popular deep learning models [29-30], [45-46] but remain to be complex when compared with classical machine learning based approaches [54], [58] which results in a higher training time. However, by parallelizing over GPUs, the inference time of the proposed end-to-end deep learning-based framework can be reduced significantly in comparison to non-parallelizable hand-engineered features that need to be extracted during the time of inference.

The idea of building "leave-one subject out", subject-independent and application agnostic EEG analysis tools is a very promising one. Even though such approaches do not outperform hand-engineered features-based EEG analysis tools, they potentially do offer a lot more generalizability over hand-engineered algorithms with respect to inter-subject variance and application agnosticism. With further efforts in this direction, it may also outperform the hand-engineered features-based EEG analysis tools. More efforts towards the explainability of such subject-independent and application agnostic EEG analysis tools is also required in the direction of applications such as source imaging, time localization and neuronal dynamics, which was not completely achieved in the proposed research work due to absence of custom collected dataset.

## 5  Conclusions

In this paper, the task of obtaining subject-invariant latent representations of multichannel EEG signals is thoroughly investigated using autoencoder based architectures. This is further followed up by the usage of these latent vectors for performing classification tasks such as emotion recognition. Attention mechanism was an essential component in all our experimented architectures due to its ability to keep track of long-term dependencies in sequential data, which cannot be modelled by LSTM or CNN due to enormous length of EEG data. We performed subject-independent emotion recognition to show that the proposed methodology was able to generalize over the whole dataset i.e., SEED and DEAP Datasets, where competitive results were achieved in comparison to the previous state-of-the-art works in the subject-independent emotion recognition domain. The comprehensive scope of the proposed framework as an application agnostic EEG analysis tool is further demonstrated on a more immediate real-life application of epileptic seizure detection and key time frame identification using the CHB-MIT dataset, where we achieved state-of-the-art results. This shows that the proposed attention neural network-based methodology can be a promising approach for diagnosing and further helping people with diseases like Epilepsy and Alzheimer, where basic functional brain networks may have been disrupted. The efficacy of the proposed approach is further employed to localize the emotional burst in time and for finding common latent vector subspaces irrespective of subjectivity of EEG data, which opens a new paradigm for studying neuronal dynamics using EEG. Thus, with this research work we have taken a step towards developing application agnostic EEG analysis tools and aim to build more comprehensive and efficient tools in the future.

**CRediT Authorship Contribution Statement**

**Arjun:** Conceptualization, Data curation, Methodology, Software, Writing - original draft. **Aniket Singh Rajpoot:** Conceptualization, Data curation, Methodology, Software, Writing - original draft. **Mahesh Raveendranatha Panicker:** Conceptualization, Supervision, Writing – review and editing.

**Declaration of Competing Interest**

The authors declare that they have no financial or personal relationships with other people or organization that could inappropriately influence (bias) their work.

*Data Availability*

Three data sets are supporting the results of this project. The DEAP dataset can be obtained from https://www.eecs.qmul.ac.uk/mmv/datasets/deap/. The SEED dataset can be obtained from https://bcmi.sjtu.edu.cn/home/seed/. The CHB-MIT dataset can be obtained from https://physionet.org/content/chbmit/1.0.0/.

*Ethical Approval*

Not required.

*Supplementary Materials*

The codes are available on request at https://github.com/arjunsinghrathore/Subject-Independent-Emotion-Recognition and will be made available public upon acceptance.

*Acknowledgement*

Authors would like to acknowledge the Indian Institute of Technology Palakkad and Ministry of Education, Government of India for their support.

# Supplementary Material

## A. More Encoder & Classifier Architecture's on DEAP & SEED Dataset

TABLE S1
MORE VALIDATION RESULTS FOR DEAP DATASET WITH DIFFERENT ENCODR AND CLASSIFER ARCHITECTURES

| CLASSIFER | ENCODER | Autoencoder (AE) with Attn | CNN-LSTM AE with Attn | CNN-LSTM AE | CNN AE with Attn | CNN AE | LSTM AE |
|---|---|---|---|---|---|---|---|
| CNN with Attn | Valence | 63.1 | 64.9 | 63.5 | 63 | 61.7 | 63.9 |
|  | Arousal | 66 | 67.7 | 66.9 | 65.1 | 63.3 | 66.5 |
| LSTM | Valence | 60.1 | 63.1 | 60.3 | 59.7 | 57.5 | 61.2 |
|  | Arousal | 62.5 | 64.6 | 61.1 | 60.2 | 60.6 | 61.7 |
| LSTM with Attn | Valence | 61.6 | 63.2 | 62 | 60.5 | 60 | 62.3 |
|  | Arousal | 63.3 | 65 | 62.4 | 61.3 | 62.1 | 62 |
| ResNet 1D | Valence | 56.1 | 57.3 | 54.2 | 55.4 | 55.1 | 56.5 |
|  | Arousal | 55.3 | 55.4 | 55.8 | 55.6 | 56.4 | 55.9 |
| Inception 1D | Valence | 56.2 | 57 | 53.2 | 56.3 | 54.8 | 57.1 |
|  | Arousal | 54.9 | 55.9 | 55.4 | 56 | 55 | 55 |
| Transformer | Valence | 58.5 | 50.5 | 53.2 | 53.7 | 54.9 | 51.6 |
|  | Arousal | 56.7 | 52.3 | 54 | 55.1 | 53.2 | 54.7 |

TABLE S2
MORE VALIDATION RESULTS FOR SEED DATASET WITH DIFFERENT ENCODR AND CLASSIFER ARCHITECTURES

| CLASSIFER | ENCODER | Autoencoder (AE) with Attn | CNN-LSTM AE with Attn | CNN-LSTM AE | CNN AE with Attn | CNN AE | LSTM AE |
|---|---|---|---|---|---|---|---|
| CNN With Attn |  | 73.5 | 75.1 | 72.7 | 71 | 70.9 | 74.2 |
| LSTM |  | 62 | 66.7 | 64.2 | 65.6 | 64.7 | 65.7 |
| LSTM with Attn |  | 65.1 | 68 | 65.5 | 69.2 | 67.7 | 68.1 |
| ResNet 1D |  | 59.9 | 59.3 | 60 | 57.1 | 58.5 | 57.3 |
| Inception 1D |  | 57 | 60.1 | 58.6 | 56.9 | 62.2 | 60 |
| Transformer |  | 59.1 | 55 | 58.7 | 57.5 | 56.4 | 54.2 |

## B. Results of Different Latent Dimensions Encoding on DEAP & SEED Dataset

TABLE S3
RESULTS FOR DEAP DATASET WITH DIFFERENT LATENT DIMENSION CHOICES(32 INPUT EEG CHANNELS)

|  |  | LSTM Autoencoder with Attention | | |
|---|---|---|---|---|
|  |  | 4 Latent Dimensions | 8 Latent Dimensions | 16 Latent Dimensions |
| CNN with Attn | Valence | 62.7 | 65.9 | 66.1 |
|  | Arousal | 65 | 69.5 | 68.8 |

TABLE S4
RESULTS FOR SEED DATASET WITH DIFFERENT LATENT DIMENSION CHOICES(62 INPUT EEG CHANNELS)

|  | LSTM Autoencoder with Attention | | |
|---|---|---|---|
|  | 8 Latent Dimensions | 16 Latent Dimensions | 32 Latent Dimensions |
| CNN with Attn | 71.9 | 76.7 | 77 |

As there is not much of a performance difference between 16 and 32 latent dimension choices for the SEED dataset, we decided to go ahead with 16 latent dimensions LSTM with attention autoencoder due lower architectural complexity in terms of parameters.

## C. Architecture Design of the Proposed Architecture

- **DEAP DATASET**

**TABLE S5**
**ARCHITECURE OF LSTM WITH CHANNEL ATTENTION AUTOENCODER FOR DEAP DATASET (14K TRAINABLE PARAMETERS)**

| Layer | Input Shape | Output Shape |
|---|---|---|
| Long-Short-Term-Memory | 8064x32 (INPUT EEG SIGNAL) | 8064x8 (LATENT ENCODINGS) |
| Attention | 8064x8 (LATENT ENCODINGS) | 8064x8 (LATENT ENCODINGS) |
| Long-Short-Term-Memory | 8064x8 (LATENT ENCODINGS) | 8064x32 (RECONSTRUCTED EEG SIGNAL) |

**TABLE S6**
**ARCHITECURE OF CNN WITH ATTENTION CLASSIFIER FOR DEAP DATASET (~3.36M TRAINABLE PARAMETERS)**

| Layer | Input Shape | Output Shape |
|---|---|---|
| **BLOCK-1**<br>Conv-1D(128 Filters, 5 Kernel Size)<br>Instance Normalization<br>PReLU<br>Dropout(0.2)<br>MaxPool(Pool Size : 2) | 8064x8 (LATENT ENCODINGS FROM LSTM WITH CHANNEL ATTENTION AUTOENCODER) | 4032x128 (BLOCK-1 FEATURES) |
| **BLOCK-2**<br>Conv-1D(256 Filters, 11 Kernel Size)<br>Instance Normalization<br>PReLU<br>Dropout(0.2)<br>MaxPool(Pool Size : 2) | 8064x8 (LATENT ENCODINGS FROM LSTM WITH CHANNEL ATTENTION AUTOENCODER) | 4032x256 (BLOCK-2 CONV-1 FEATURES) |
| Conv-1D(512 Filters, 11 Kernel Size)<br>Instance Normalization<br>PReLU<br>Dropout(0.2) | 4032x256 (BLOCK-2 CONV-1 FEATURES) | 4032x512 (BLOCK-2 CONV-2 FEATURES) |
| Concat | Concatenate(<br>[BLOCK-1 FEATURES(Shape : 4032x128),<br>BLOCK-2 CONV-2 FEATURES(Shape : 4032x512)] | 4032x640 (CONCAT FEATURES) |
| Highway Network | 4032x640 (CONCAT FEATURES) | 2016x640 (HIGHWAY FEATURES) |
| Attention | 2016x640 (HIGHWAY FEATURES) | 2016x320 (ATTENTION FEATURES) |
| Linear Dense<br>Instance Normalization | 2016x320 (ATTENTION FEATURES) | 2016x256 (DENSE FEATURES) |
| Flatten | 2016x256 (DENSE FEATURES) | 516,096 (FLATTEN FEATURES) |
| Softmax Dense | 516,096 (FLATTEN FEATURES) | OUTPUT CLASSES |

- **SEED DATSET**

**TABLE S7**
**ARCHITECURE OF LSTM WITH CHANNEL ATTENTION AUTOENCODER FOR SEED DATASET (54K TRAINABLE PARAMETERS)**

| Layer | Input Shape | Output Shape |
|---|---|---|
| Long-Short-Term-Memory | 30000x62 | 30000x16 |
| Attention | 30000x16 | 30000x16 |
| Long-Short-Term-Memory | 30000x16 | 30000x62 |

**TABLE S8**
**ARCHITECURE OF CNN WITH ATTENTION CLASSIFIER FOR SEED DATASET (~6.2M TRAINABLE PARAMETERS)**

| Layer | Input Shape | Output Shape |
|---|---|---|
| ***BLOCK-1***<br>Conv-1D(128 Filters, 5 Kernel Size)<br>Instance Normalization<br>PReLU<br>Dropout(0.2)<br>MaxPool(Pool Size : 2) | 30000X16<br>(LATENT ENCODINGS FROM LSTM WITH CHANNEL ATTENTION AUTOENCODER) | 15000x128<br>(BLOCK-1 FEATURES) |
| ***BLOCK-2***<br>Conv-1D(256 Filters, 11 Kernel Size)<br>Instance Normalization<br>PReLU<br>Dropout(0.2)<br>MaxPool(Pool Size : 2) | 30000X16<br>(LATENT ENCODINGS FROM LSTM WITH CHANNEL ATTENTION AUTOENCODER) | 15000x256<br>(BLOCK-2 CONV-1 FEATURES) |
| Conv-1D(512 Filters, 11 Kernel Size)<br>Instance Normalization<br>PReLU<br>Dropout(0.2) | 15000x256<br>(BLOCK-2 CONV-1 FEATURES) | 15000x512<br>(BLOCK-2 CONV-2 FEATURES) |
| Concat | Concatenate(<br>[BLOCK-1 FEATURES(Shape : 15000x128),<br>BLOCK-2 CONV-2 FEATURES(Shape : 15000x512)] | 15000x640<br>(CONCAT FEATURES) |
| Highway Network | 15000x640<br>(CONCAT FEATURES) | 7500x640<br>(HIGHWAY FEATURES) |
| Attention | 7500x640<br>(HIGHWAY FEATURES) | 7500x320<br>(ATTENTION FEATURES) |
| Linear Dense<br>Instance Normalization | 7500x320<br>(ATTENTION FEATURES) | 7500x256<br>(DENSE FEATURES) |
| Flatten | 7500x256<br>(DENSE FEATURES) | 1,920,000<br>(FLATTEN FEATURES) |
| Softmax Dense | 1,920,000<br>(FLATTEN FEATURES) | OUTPUT CLASSES |